**Psychosis involves a deficit of information compression in connected speech**

Samuele Vallisa[1,*], Claudio Palominos[1], Rui He[1], Emre Bora[2,4], Burcu Verim[2], Cemal Demirlek[3], Berna Yalincetin[2], Philipp Homan[5,6], Wolfram Hinzen[1,7]

[1] Grammar and Cognition Lab, Department of Translation & Language Sciences, Universitat Pompeu Fabra, Barcelona, Spain
[2] Department of Neurosciences, Health Sciences Institute, Dokuz Eylul University, Izmir, Turkey.
[3] Department of Psychiatry, McLean Hospital, Harvard Medical School, Belmont, Massachusetts, United States of America.
[4] Department of Psychiatry, Dokuz Eylul University, Izmir, Turkey.
[5] Department of Adult Psychiatry and Psychotherapy, University of Zurich, Zurich, Switzerland.
[6] Neuroscience Center Zurich, University of Zurich and ETH Zurich, Zurich, Switzerland.
[7] Institut Català de Recerca i Estudis Avançats (ICREA), Barcelona, Spain

[*] **Correspondence to**: Samuele Vallisa, Grammar and Cognition Lab, Department of Translation & Language Sciences, Universitat Pompeu Fabra, Carrer de Roc Boronat 138, Sant Martí, 08018 Barcelona, Spain email: samuele.vallisa@upf.edu.

## Abstract

Large language models (LLMs) with human-like performance on linguistic tasks have transformed the study of language in neurodiverse conditions. LLMs provide representations of linguistic input in the form of high-dimensional vectors (embeddings), and next-token predictions computed from these embeddings. Previous crosslinguistic evidence suggests a complexity reduction in the form of both lower intrinsic dimensionality (ID) of LLM representations and higher mean surprisal (prediction error) in psychosis. We hypothesized that these metrics reflect a general deficit of information compression in psychosis, linked to grammatical organization as what enables predictions in language.We operationalized surprisal difference as the difference between surprisal as estimated from word frequency and surprisal as based on a contextual LM, which is sensitive to grammatical organization over and above lexical concepts. Using a dataset of 144 Turkish speakers, including 106 patients with schizophrenia-spectrum disorders (SSD)—56 with chronic schizophrenia (SZH), 33 with first-episode psychosis (FEP), and 17 with schizoaffective disorder (SZA)—and 38 healthy controls. We report: (1) Surprisal difference is attenuated in all clinical groups relative to controls, independently of word count; (2) Compressibility (intrinsic dimension) is reduced in SZH and FEP; (3) Syntactic complexity and compressibility both predict surprisal difference. These results, further refining an alteration in the geometry of the semantic space in psychosis as previously attested, suggest a broader deficit in information compression in this disorder, with a mechanistic underpinning in the operations of grammar.

## Introduction

Embeddings and predictions from large language models (LLMs) have become critical windows into clinical alterations of the informational dynamics of speech[1]. LLMs provide representations of linguistic input in the form of high-dimensional vectors (embeddings), from which predictions for next tokens are computed by the model. In psychosis, spontaneous speech shows an increase in surprisal (prediction error),[2] which relates to 'loose associations' (formal thought disorder),[3,4] and to positive and general symptom items as assessed with the PANSS.[2] Sharpe et al.[4,5] specifically argue for an attenuated context-based facilitation of prediction in psychosis, leading to lesser reduction of surprisal through context. Psychosis also shows a cross-linguistically corroborated increased compression ('shrinkage') of the lexical-semantic space, as manifest through higher mean cosine similarity between word embeddings[2,6–8], smaller hypervolumes of the convex hull enclosing these embeddings [9], and lower intrinsic dimensionality (ID)[10]. In information theory as well as empirically, prediction and representation are connected: surprisal measures how much a linguistic representation resists compression under an ideal predictive code, with highly expected inputs being more compressible and unexpected inputs less compressible [1112]. We therefore here aimed to target compressibility in psychosis in an integrated analytical scheme.

This was based on the foundational concept that information compression is a necessity in language processing and grammatical organization is a crucial factor in making language more predictable. Intuitively, next words are essentially unpredictable in arbitrary word lists, but relatively predictable in sentences. Surprisal as based on probabilities from word frequencies is thus expected to be higher than when based on grammatical context. As LLMs are sensitive to grammar, we operationalized surprisal difference as the difference between surprisal as estimated from word frequency and surprisal from contextual LLM[13]. We raised four interrelated questions: (1) Whether surprisal reduces when moving from a non-contextual to a contextual predictive model (2) Whether grammatical complexity predicts surprisal difference; (3) Whether ID is reduced in psychosis; and (4) Whether surprisal difference and ID are related. Evidence of a context-based predictive coding deficit in psychosis[14,15] predicts attenuated surprisal difference in this population, which we also hypothesized to be related to greater compression in psychosis as measured with ID, based on empirical evidence[16] of this compression in psychosis and theoretical evidence[12].

## Methods

### *Participants*

Participants were recruited through the Psychotic Disorders Outpatient Unit of the Department of Psychiatry, Dokuz Eylul University, Izmir. Advertisements at the university campus and at the university hospital were used to include 38 healthy controls (HCs) who had no family or personal history of mental illness. Every participant spoke Turkish as their first language. Participants were excluded if they had a personal history of medical or neurological disorders or if they were currently abusing alcohol or other substances. The Structured Clinical Interview for DSM-IV Axis I Disorders was used. The patient group (n=106 with schizophrenia spectrum disorders, SSD) comprised chronic schizophrenia (SZH), schizoaffective disorder (SZA) and first episode psychosis (FEP). The dataset includes transcriptions of speech obtained from three different tasks: picture description (PD), free speech (FS) and retelling (RT). For PD, three different images were used. The free speech task includes conversational speech about participant's past, personal narrative, health narrative. The clinical scores collected include information regarding positive symptoms, negative symptoms and social performances.

Table 1 summarizes the sample including basic demographic and clinical characteristics as well as total word count. As revealed by a post-hoc pairwise Dunn's test with Bonferroni correction, the effect of word count was driven by SZH, who produced less words than HC ($p<0.01$). As a further measure of lexical diversity, we also computed the moving average type-token ratio (MATTR) on a sliding window of 50 words across the texts with a sliding parameter of 12 words (see group results in Table 1). The complete TTR analysis is presented in the Supplementary Materials (SM).

All linguistic features described in the following sections (surprisal, syntactic metrics, and dimensionality) were computed per participant per task. Responses were first concatenated into a single text, then segmented into sentences, which served as the unit of analysis. Metrics were calculated for each sentence and subsequently averaged across all sentences within a task. Sentences containing fewer than five words were excluded to ensure the reliability of the linguistic metrics. For metrics that produce a value at the word level, scores were first averaged within each sentence before being aggregated by task.

*Ethical aspects*

The data collection protocol was approved by the Dokuz Eylül University Non-Interventional Research Ethics Committee (File/Protocol No: 8527-GOA). Written informed consent of all participants was obtained before enrolling them in the data collection.

*Linguistic metrics: Surprisal*

The surprisal of a sentence $s$ was defined as:

$$Surprisal(s) = -\sum_i log_2 p(w_i) \qquad (1)$$

where $p(w_i)$ represents the measured probability of a certain word $w_i$ belonging to $s$. We distinguished between two sources of surprisal: (i) Frequency surprisal, which is independent of context, obtained by assigning to each word a probability based on the overall word frequency. For frequencies we used the wordfreq[17] python library, and from the frequencies we computed a probability of occurrence, which can be fed into equation (1). (ii) BERT surprisal, which is computed taking the sentence context into account. We first extracted the probabilities that the model (dbmdz/bert-base-turkish-uncased1) assigned to each word given the remaining sentence context. Given a sentence composed by the words [$w_1$, $w_2$, $w_3$...], we iteratively masked one word at a time and fed the partially masked sentence into the model. BERT then outputs a probability distribution over its vocabulary words (about 32,000 words) for the masked word. For each sentence, we fed these word probabilities into equation (1) to obtain a single value for each sentence. We defined *surprisal difference* as the difference between the frequency-based and BERT-based surprisals. This measure captures how contextual information influences the baseline surprisal, marking the transition from a grammar-free to a grammar-sensitive surprisal.

*Syntactic metrics*

To study the relation between surprisal difference and syntactical complexity, we used both dependency grammar and constituency theoretic frameworks. Dependency grammar is often used to obtain dependency trees, which reveal the structure of an expression in terms of hierarchical links [5]. These trees are typically directed, where vertices are words and links indicate syntactic dependencies between words. In dependency trees, the distance between nodes (i.e., words) can be quantified by the number of words separating them,[18,19] yielding a notion of dependency distance as the linear distance between two hierarchically related words in a dependency tree. On the other hand, constituency trees are designed to represent the hierarchical relations between phrasal constituents of a sentence, such as noun phrases and verb phrases. Hierarchical distance was defined as the number of edges on the path from the root node of the tree to that node in the constituency tree of a sentence. After extracting the dependency and hierarchical distances of each sentence, we computed two metrics: (i) Dependency Optimality (**Ω**), which quantifies the extent to which dependency distances are optimized relative to both a random (uniform) arrangement of words and the configuration that minimizes those distances.[19] It is defined as:

$$\Omega = \frac{D_{rla} - D}{D_{rla} - D_{min}} \qquad (2)$$

where $D$ are the observed dependency distances, $D_{rla}$ is the average value of $D$ over all possible linear arrangements and $D_{min}$ is the minimum distance observable given the dependencies relations. (ii) *Mean hierarchical distance* (MHD), which is the average hierarchical distance across all words in the sentence. A lower MHD represents a flatter structure, where words are on average close to the root, while a higher MHD stands for a nested, thus more complex, structure. Constituency parsing, on the other hand, is used to extract the syntactical structure of a given sentence and the sub-sentence nested in it. We extracted constituency trees to evaluate: (iii) Mean constituency depth (MCD), which captures the vertical complexity of a sentence's constituency syntactic tree. The MCD of a sentence $s$ is obtained as:

$$MCD(s) = \frac{1}{n}\sum d(w_i)$$

where $n$ is the number of words, and $d(w_i)$ identifies the path length from the root node to the word $w_i$. A higher MCD indicates a more nested sentence, which indicates a higher usage of subordinate clauses and more complex phrase structures.

*Embeddings*

We obtained word embeddings from BERT[20], a pre-trained deep language model that provides contextual word embeddings, specifically *dbmdz/bert-base-turkish-uncased*, the base BERT model for Turkish language, which produces 768-dimensional embeddings. Sentences were given as inputs to the model, which operates by masking one word at time and making use of the remaining context to produce embeddings. For each input word, thus, BERT provides an embedding vector capturing semantic and syntactic information obtained by the context.

*Intrinsic dimensionality (ID)*

ID identifies the minimum number of variables required to describe a system within a specific approximation error range, using linear or nonlinear methods[21]. It is used to extract meaningful information from high dimensional data. Within the high-dimensional space they live in, these data often lie on curved manifolds of a much lower dimension. In other words, the number of dimensions required to effectively represent their structure is much smaller. To compute a text's ID we used a nearest-neighbour based approach. Nearest-neighbour methods are used to find the geometrically closest points to another given point[22,23]. They unveil local properties of the system: the statistical distribution of distances between neighbouring tokens, capturing proximity at the local scale. The transition to a global description is achieved using a statistical estimator that aggregates the nearest-neighbour distances information. The fundamental idea is estimating the ID by using the ratios $\mu_{i,n_1,n_2} = \frac{r_{i,n_2}}{r_{i,n_1}}$ where $r_{i,k}$ is the distance between point *i* and its *k-th* nearest neighbour and $1 \leq n_1 \leq n_2$. The estimator choice follows the assumption of a unique manifold, which allows the aggregation of these disjoint local estimates into a global value[24,25]. In this work, we use different types of estimators.

The first is Two Nearest Neighbours[23] (2NN), which determines the ID by utilizing, for each point, the distances from the first and second closest data points[23]. Since this method consider only the first two neighbours, it is related to very local proprieties of the manifold. Moreover, this method is less dependent on smooth data density variations. We used the algorithm implemented in the python library *scikit-dimension*[26]. The second estimator is *Generalised ratios id estimator* (Gride)[27,28], which generalizes the 2NN estimator. Gride computes ID at different scales (i.e. at different ratios): by using this algorithm the above parameter $n_2$, known as the range scaling parameter, is varied before returning the estimated value[27]. We used two different approaches to select the scaling parameter: 1) *kstar* Gride, which was used for most of the analysis and is based on an automatic estimation of the parameter $k$, i.e. the neighbour index. For each point of the set this parameter is computed by using a statistical test, which aims to find the largest neighbourhood in which the density can still be considered relatively constant[29]. This approach is useful when, as in our case, the manifolds have a large variety of point density. Moreover, by maximizing the neighbourhood, it provides a more global view with respect to the 2NN algorithm. 2) The classical Gride algorithm varying the range scale manually, which was used to study how ID varies across different scales. Both approaches are implemented in the python library *DADApy*[29]. Finally, we use the MLE estimator. This algorithm operates by assuming that data points are locally sampled from a uniform distribution, and the distances to the $k$-th nearest neighbours are modelled as a non-homogeneous Poisson process. The ID is then derived by maximizing the log-likelihood function of these observed distances. For all our main analysis and comparisons, we used the 2NN and the kstar Gride methods, while, only for the specific scales analysis presented in section *Relationship between surprisal difference and dimensionality*, we used the classical Gride and MLE algorithms.

*Post-hoc analysis*

In light of the significant positive effect of word count, we added a post-hoc sliding-window analysis of surprisal difference. We computed surprisal difference across a range of window sizes spanning from 50 to 300 words. We also set a varying stride parameter, which represents the step size, where the size is defined as the number of words by which the window advances for each successive calculation. To maintain a consistent degree of data overlap across scales, the stride for each window was defined as 25% of the window size, effectively displacing each subsequent analysis by a quarter of the window length for each given size.

*Statistical Analysis*

To account for the non-independence of observations resulting from multiple linguistic samples per participant, and to isolate group effects, we employed General Estimating Equations (GEE). All continuous predictors and continuous covariates were standardized (using z-score standardization) prior to analysis to ensure that the resulting beta coefficients represent effect sizes in units of standard deviation, facilitating direct comparison between metrics. For each dependent variable we selected, the model was specified as follows:

$$y_{ij} = \beta_0 + \sum_{k=1}^{p} \beta_k X_{kij} + \epsilon_{ij}$$

where $y_{ij}$ represents the outcome for participant $i$ at observation $j$, $X_{kij}$ represents the set of $p$ predictors, $\beta_0$ the model intercept, $\beta_k$ the regression coefficients and $\epsilon_{ij}$ the residual error term. These models were used to examine group differences in the TTR (see Figure S2), in surprisal difference and in ID, using HC as baseline and controlling for potential confounders, namely age, sex, education, and number of words. A similar statistical framework was applied to assess the relationship between syntactic complexity metrics and surprisal measures, and lexical diversity and ID. In this case we did not investigate group differences but evaluated the syntactic metrics as predictors for surprisal difference, adjusting for the same confounders as above. We used GEEs to evaluate the contribution of ID and the interaction between diagnosis and ID as predictors of surprisal difference. The complete formulas of GEE regressions are provided in Supplementary Materials. Finally, to assess mapping onto clinical measures, we calculated the Pearson correlations between surprisal and dimensionality variables with clinical and behavioural indices.

## Results

### *Surprisal*

GEEs models confirmed diagnosis as a significant predictor of BERT surprisal, while showing no significant effects for frequency surprisal. In line with our key prediction, surprisal difference was significantly attenuated in SZH, SZA and FEP groups, compared to HC (Figure 1). In the windowed post-hoc analysis, average surprisal difference remained significantly lower in the clinical population, for every window size (Figure 1b), suggesting robustness of this effect against speech quantity.

We also confirmed our prediction of a predictive relationship between grammatical complexity and surprisal metrics (Figure 1c). Weak effects were found for Omega, while the strongest effects were observed for MHD and mean depth, especially in the case of surprisal difference, where all coefficients were positive, and for BERT surprisal, where the relationship was negative. Effects on frequency-based surprisal were weaker.

### *Intrinsic dimensionality*

In line with our predictions, the Gride kstar analysis revealed significant standardized negative effects of diagnosis on ID for both FEP ($\beta \approx -1.01$) and SZH ($\beta \approx -0.91$), alongside a strong positive influence of the number of words ($\beta \approx +2.17$) (Figure 2a). The same groups showed a negative effect when using the 2NN method ($\beta_{SZH} \approx -0.57, \beta_{FEP} \approx -0.56$) ,but the group effect of FEP was not significant in this case. Moreover, compared to the Gride method, the diagnostic group effects in this case were reduced in magnitude (Figure 2b). Finally, we evaluated the relationship between ID and three distinct measures of lexical diversity. Specifically, in addition to the previously described MATTR, we computed a lemmatized version to account for morphological variation and a content-word-only version and isolate semantic density from functional grammatical markers. Following the same statistical framework as applied to the study of syntactical metrics, we utilized GEE to evaluate these metrics as predictors of ID. The results revealed consistent positive effects across all diversity metrics in predicting ID, though the magnitude of these effects was notably attenuated when using the Gride kstar estimator compared to other scales (Figure 2c).

*Relationship between surprisal difference and dimensionality*

We fitted another GEE with surprisal difference as the dependent variable, using diagnosis, ID, and their interaction as predictors. We found two significant but opposite effects of ID on surprisal difference for the two different methods: a positive effect between ID and surprisal difference for the Gride kstar method (Figure 3a, $\beta \approx +0.27$, $p<0.001$), and a negative one for the 2NN method (Figure 3b, $\beta \approx -0.45$, $p<0.001$). As this difference could be related to the different locality properties of the two methods, we conducted an analysis of the IDs varying the neighborhood scale, defined as the number of nearest neighbors ($k$) used to define the local geometry around each data point, transitioning from a small scale ($k=2$), capturing micro-topology, to larger scales (up to k=256) capturing the meso-scale. Figure 3c shows that at the smallest scales the effect of ID on surprisal difference is negative, while it becomes positive at larger ones. Interestingly, in the 2NN case we found a significant attenuation of the effect for the interaction term SZH*ID (Figure 3b, $\beta \approx +0.31$, $p=0.003$). This suggests that the impact of the ID predictor on surprisal is significantly reduced in SSD when using 2NN.

*Relationship with clinical scores*

Higher surprisal difference was significantly associated with higher SCIP scores, especially with working memory and processing speed. The opposite relation was found for BERT surprisal, which was also negatively associated with the SAPS delusions score. ID Gride was significantly and negatively associated with BNSS expressive capacity, while demonstrating robust positive correlations with cognitive performance, particularly in verbal fluency and processing speed. By contrast, higher ID 2NN was associated with significantly lower scores for SAPS bizarre behaviour, SAPS positive formal thought disorder, and BNSS amotivation, while lacking significant correlations with cognitive scores.

## Discussion

This study aimed to connect representation and prediction as two crucial dimensions of the informational dynamics of speech as measured with LLMs. Overall, the results confirmed our broad hypothesis of a deficit of information compression in psychosis, consistent with previous evidence of both lower ID of the embedding space and increased contextual surprisal in psychosis[16,30], while empirically adding insight through a correlation between these two. Our GEE models controlling for word count and our windowed analyses together support the conclusion that this deficit in information-theoretic compressibility is robust to a difference in speech quantity often documented in psychosis and found empirically in one of our clinical groups as well, relative to controls. The same deficit is consistent with a consistently lower TTR in our clinical groups, as the TTR itself can be seen as an additional (model-free) index of compressibility of the lexical semantic space[31]. Lesser lexical variability (sampling from the lexical-conceptual semantic space) likely shapes the geometry of the embedding manifold, and the correlations between both TTR and ID with surprisal difference suggest that linguistic redundancy is significantly linked to probabilistic compression. Interestingly, the effect of lower ID in psychosis was consistent across SZH and FEP, while not reaching significance for SZA, the least severe among the groups in our sample in terms of symptoms.

Information compression as such leaves open the question of the mechanism accomplishing such compression. Today's LLMs are the best available mechanistic approximations of language as a system. LLM representations incorporate both information about the overall word frequency[32] and about context involving grammatical organisation. Expressing a coherent thought that is related to the world due its content requires grammar, and a grammatically complete configuration (a sentence) always expresses a thought in this sense. Predicting words in grammatical contexts is thus linked to predicting what thought is being expressed. In turn, grammatical meaning shapes lexical choice in production, where we choose the words in line with the higher-level meanings (thoughts) we wish to express, and which correspond, not to lexical concepts, but to concepts in specific relationships that are mirrored in grammar. In this respect, a crucial finding in our study is the *lack* of significant group differences in frequency-based surprisal, pointing towards a predictive processing deficit being present in psychosis only when grammatical organization is considered. In line with this idea, surprisal difference associated with our independent grammaticality complexity measures. This is consistent with evidence in He et al.[13] that surprisal difference, when moving from frequency-based to context-based surprisal (*adding* grammar), was highly correlated to the effect of moving from contextual surprisal in reversed sentences to contextual surprisal in normal sentences (erasing grammar). It is also consistent with an attenuated context-based facilitation of prediction in psychosis documented in Sharpe et al.[4,5], though it gives this notion a specifically grammatical spin. However, our results regarding grammatical complexity are correlational and our methodology cannot exclude that the mechanism behind uncertainty reduction in language is not linguistic (or grammar) per se, but located in some language-independent factor.

Model differences modulating the relationship between surprisal and ID, specifically between Gride and 2NN, show the complexity of this relation. Insights on the properties and relations between dimensionality reduction algorithms are an active area of discussion. Specifically, two recent studies[12,33] provide opposite conclusions on the relationship between contextual surprisal and ID. To remain methodologically more agnostic, we employed a range of ID estimation methods across varying scales. While the inherent variability of our dataset and limited data quantity likely influences the outcomes, it is striking that a shift in the association between ID and surprisal difference emerges when shifting from more local to more global ID computation methods (Figure 3c).

Notably, we observed that distinct ID estimators correlate with different symptom dimensions within the clinical population. While the local estimator (ID 2NN) significantly correlated with both positive and negative symptoms, the more global estimator (ID Gride kstar) showed a robust association with cognitive performance. This divergence suggests that different topological properties of the linguistic manifold, captured at varying scales, represent distinct aspects of the discourse in psychosis. Our tentative interpretation is that 2NN, by investigating local aspects of the discourse, intercepts more semantically related features of it. Specifically, 2NN might be sensitive to word clusters or scatter of words internally in those clusters, while ID Gride captures global aspects of the manifold geometry. It thus seems reasonable that the latter method aligns more robustly with cognitive performance, such as processing speed and verbal fluency, which represent the overall complexity and organization of the discourse. We recognize the interpretation of ID across different scales is an ongoing field of research[34–36], making it challenging to definitively link these insights with specific cognitive or symptomatic scores.

Overall, our results suggest that grammar as an information compression device is a new foundational concept that merits close consideration in the context of psychosis, where a wide spectrum of linguistic variables has consistently shown group differences and predictive significance, while an overarching conceptual framework for these language alterations remains missing. Recent work, in particular, has shown that the still standard assumption that computational linguistic metrics like mean semantic (cosine) similarity measure coherence in speech, is ill-founded[37]. The present approach can contribute to such a conceptual framework, though it too remains incomplete. In particular, a reduction in uncertainty cannot be *per se* what is optimized in language, as speakers would cease to produce anything that is informationally new. Language must defy prediction as much as it depends on it. A framework truly explaining how grammar *optimizes* uncertainty is needed.

*Limitations*

While our analysis focused exclusively on participant responses, the influence of the interviewer's interruptions / interventions on the speech output and our measures is potentially significant. Clinical speech elicitation paradigms often do not prioritize long, uninterrupted speech sequences, and these are generally difficult to obtain. A second methodological challenge is that although we already used different methods to compute ID, the range of possibilities in this terrain is large, with some recent methods particularly focusing on estimations of the correct scale for computing ID[38,39]. Methods used to compute ID are also inherently sensitive to sample size, often exhibiting low stability for low amounts of datapoints (see Figure S5a). A third issue is variability across LLMs. Testing our framework across different model architectures may provide a more comprehensive view.

## Conclusions

In this study we targeted meaning uncertainty in psychosis linking a geometric and probabilistic perspective under the overarching perspective of information compression. We confirmed a compromised uncertainty reduction linked to grammar, a reduced dimensionality of the semantic embedding space, and a relationship between uncertainty and dimensionality. While this work establishes a link between the geometry of language and its predictability in psychosis, further validation is required to determine how these latent dynamics might reliably index clinically relevant parameters such as symptoms, prognosis, or severity; and fit into an integrated conceptual scheme identifying the key balance between predictability and surprisal in language.

**Data availability**

Requests to access the datasets should be directed to emre.bora@deu.edu.tr

**Author contributions**

S.V. performed experiments, analysed data and developed figures. C.P., R.H., and S.V. designed analysis procedures. W.H., R.H., C.P., and S.V. conceptualized the work. W.H. and S.V. co-wrote all paper sections. E.B., B.Y., C.D., and B.V. collected the data and performed the clinical evaluations. W.H. and P.H. supervised the study and acquired funding. All authors reviewed and approved the final manuscript.

## Funding and declaration of interest

*Funding*

This research was supported by European Research Council (ERC-2023-SyG, 101118756). Views and opinions expressed are however those of the authors only and do not necessarily reflect those of the European Union or the Agency. Neither the European Union nor the granting authority can be held responsible for them. Data acquisition was supported by the Scientific and Technological Research Council of Turkey (TUBITAK 2247 Project No: 120C141) which had no further role in the study design, in the collection, analysis and interpretation of data and in the decision to submit the article for publication.

*Declaration of interest*

P.H. has received grants and honoraria from Novartis, Lundbeck, Mepha, Janssen, Boehringer Ingelheim, OM Pharma and Neurolite outside of this work.

**Table 1**: Demographic characteristics, clinical assessment, and total word count across diagnostic groups.

| Variables | HC | SZA | FEP | SZH | Test | p-values |
|---|---|---|---|---|---|---|
| **Count** | 38 | 17 | 33 | 56 | / | / |
| **Age** | 34.92 (11.65) | 38.41 (8.82) | 24.97 (8.85) | 35.52 (10.69) | ANOVA | 0.000 |
| **Sex (female %)** | 8.78% | 5.41% | 12.16% | 14.18% | X2 test | 0.293 |
| **Education** | 13.79 (3.45) | 12.71 (3.21) | 12.57 (3.87) | 11.13 (3.02) | ANOVA | 0.004 |
| **SAPS** | / | 13.53 (11.98) | 16.52 (12.41) | 22.07 (18.31) | ANOVA | 0.032 |
| **BNSS** | / | 13.88 (6.47) | 27.00 (15.20) | 25.50 (15.63) | ANOVA | 0.004 |
| **PSP** | 86.53 (6.63) | 57.00 (10.76) | 44.90 (16.03) | 40.80 (13.22) | ANOVA | 0.000 |
| **SCIP** | 55.37 (13.83) | 59.13 (9.05) | 66.04 (12.88) | 75.71 (16.28) | ANOVA | 0.002 |
| **Word count** | 131.8 (40.03) | 123.3 (43.47) | 107.8 (60.73) | 94.9 (51.62) | Kruskal-Wallis | 0.003 |
| **MATTR** | 0.842 (0.05) | 0.845 (0.06) | 0.822 (0.07) | 0.830 (0.07) | ANOVA | 0.004 |

*Abbreviations:* HC: healthy controls, SZA: schizoaffective disorder, FEP: first episode psychosis, SZH: schizophrenia, SAPS: Scale for the Assessment of Positive Symptoms, BNSS: Brief Negative Symptom Scale, PSP: Personal and social performances, SCIP: Screen for Cognitive Impairment in Psychiatry, MATTR: Moving Average Type-Token Ratio.

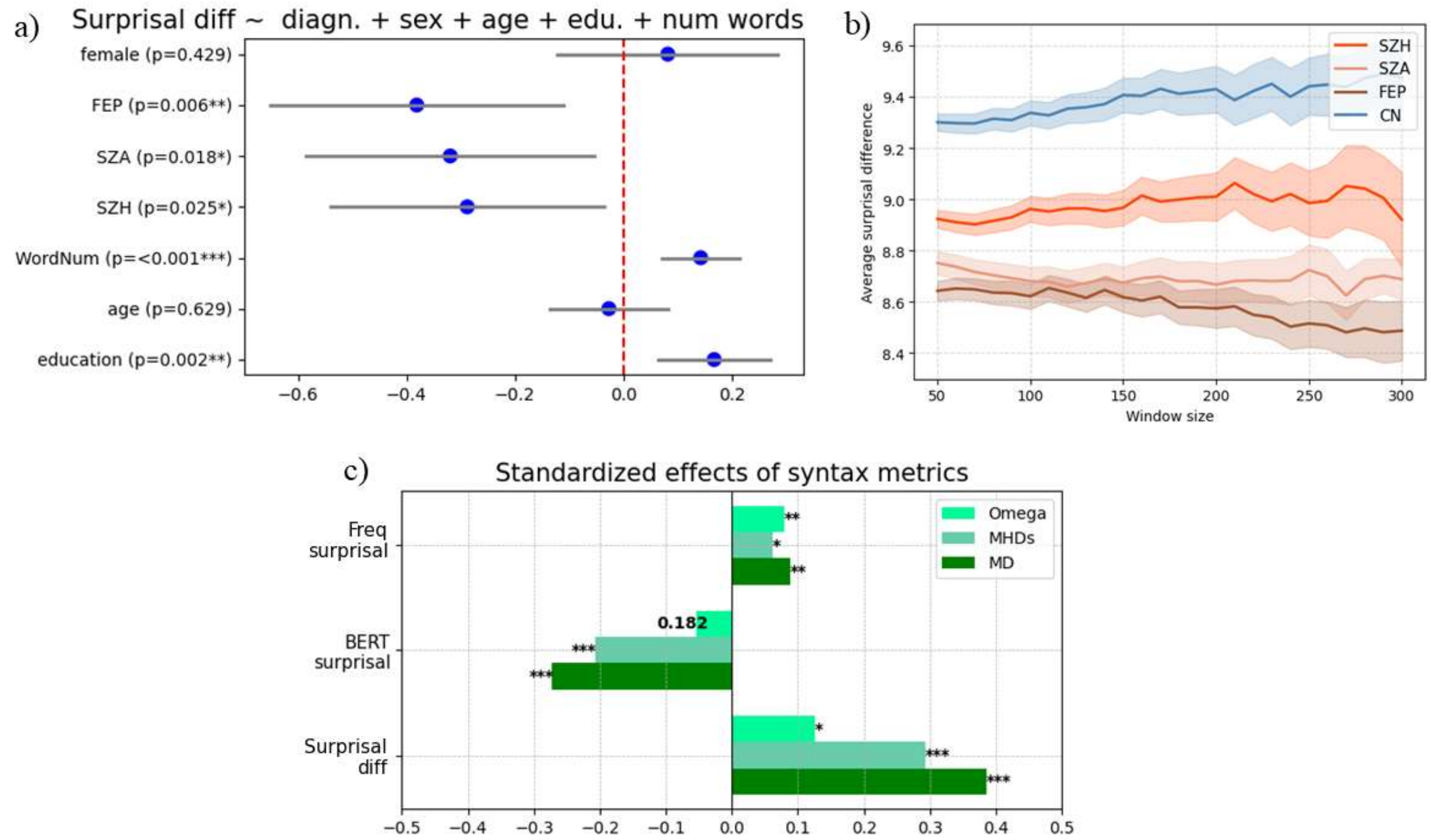

**Figure 1:** Surprisal metrics across groups. *a)*: Forest plot visualizing the standardized coefficients from a general linear model (GEE) predicting surprisal difference, controlling for gender (female), age, education, and the number of words. The vertical dashed line at 0 represents the baseline (HC). Asterisks denote statistical significance levels *(* $p < 0.05$, ** $p < 0.01$, *** $p < 0.001$)*. *b):* Average surprisal difference using a windowed analysis, across different window sizes. Solid lines represent the mean average surprisal difference; shaded bands indicate the represent ±1 standard error of the mean. *c):* Standardized GEE coefficients showing the effects of syntactical metrics (Omega, MHD, Mean depth) on frequency surprisal, BERT surprisal, and surprisal difference, adjusting for word count, age, education, and sex.

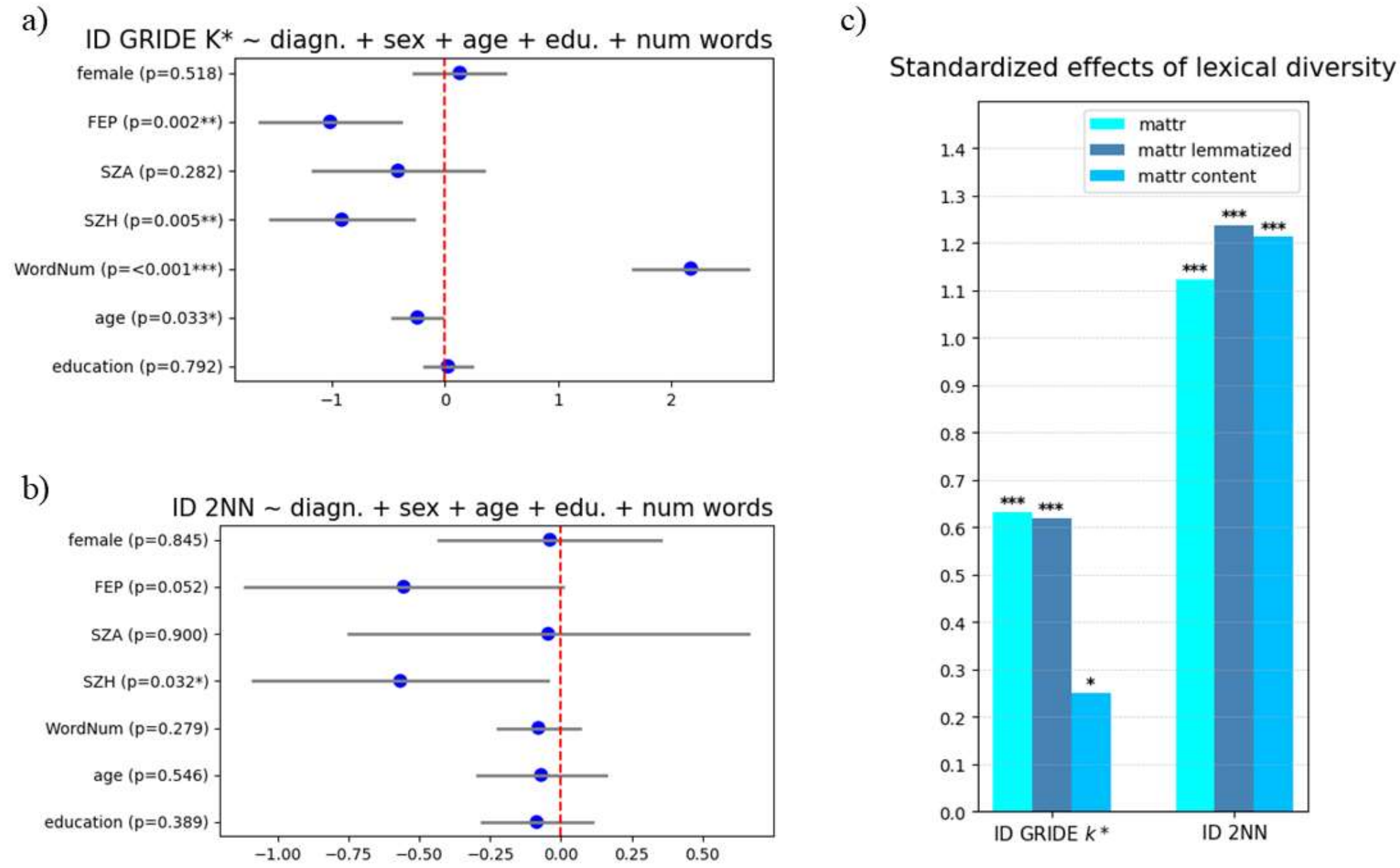

**Figure 2**: Individuals with schizophrenia and first-episode psychosis use a narrower range of meanings in their speech, and this narrowing is linked to lower lexical diversity. *a*) Forest plot visualizing the standardized coefficients from a GEE predicting ID from Gride kstar based on diagnosis, sex, age, education, and word count. The vertical dashed line at 0 represents the HC group. *b)* Forest plot showing the standardized coefficients from predicting 2NN-based ID using the same estimators. *c)* Standardized GEE coefficients showing the effects of lexical diversity metrics (MATTR, lemmatized MATTR and content MATTR) on ID 2NN and ID Gride kstar, adjusting for word count, age, education, and sex

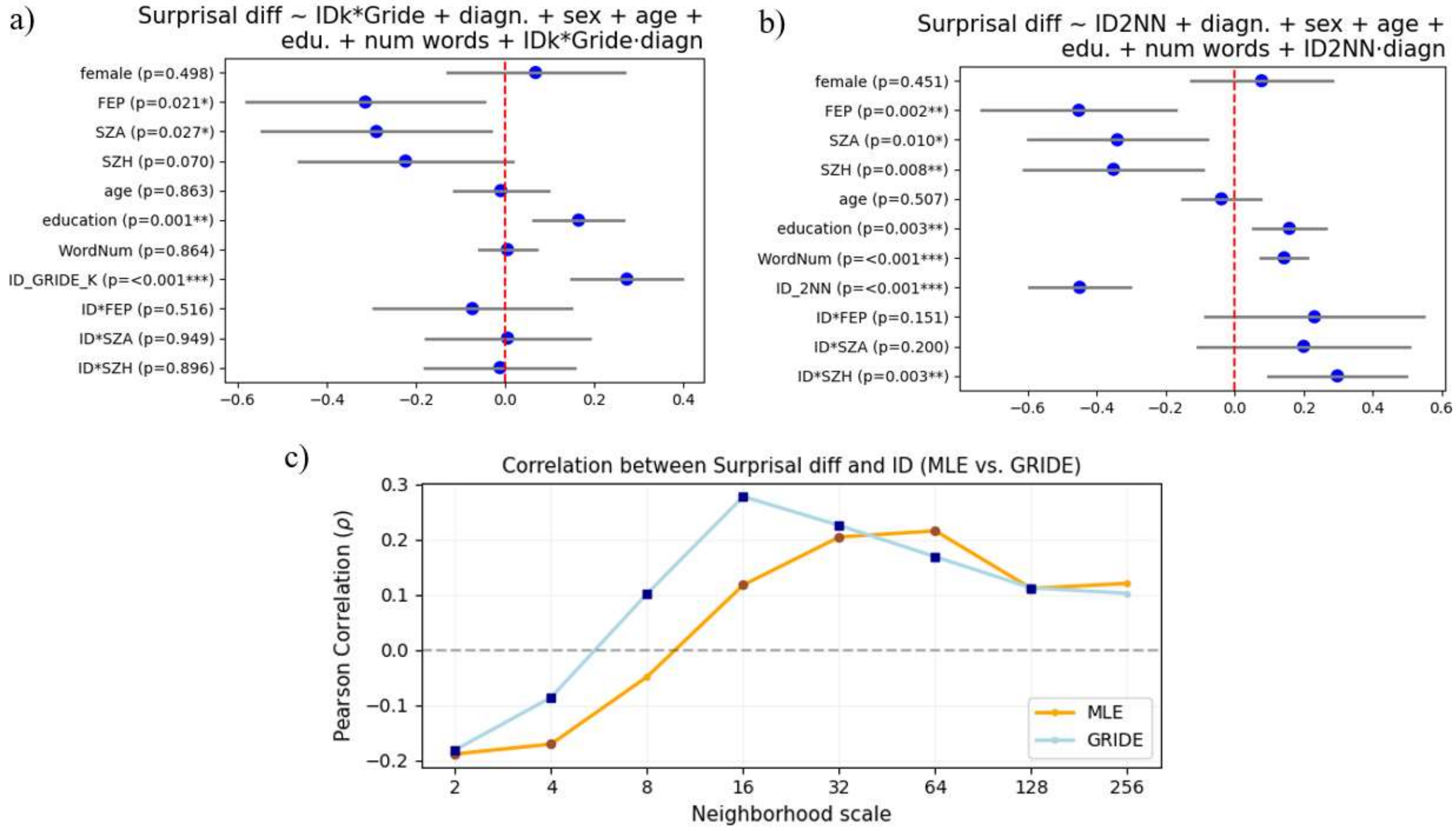

**Figure 3**: Relationship between surprisal difference and dimensionality (a-b): forests plot visualizing the standardized coefficients from a general linear model (GEE) predicting surprisal difference from ID as calculated with kstar Gride (a) and 2NN (b) as dimensionality estimators, controlling for gender (female), age, education, and the number of words. The vertical dashed line at 0 represents the baseline (HC group). Asterisks denote statistical significance levels (* $p < 0.05$, ** $p < 0.01$, *** $p < 0.001$). c) Correlation analysis showing the shift in the Pearson correlation coefficient between surprisal difference and ID across different neighbourhood scales for both MLE (orange) and Gride (light blue). The x-axis represents the number of neighbours considered in a log-scale. Significant correlations ($p < 0.05$) are indicated by larger dots or squares.

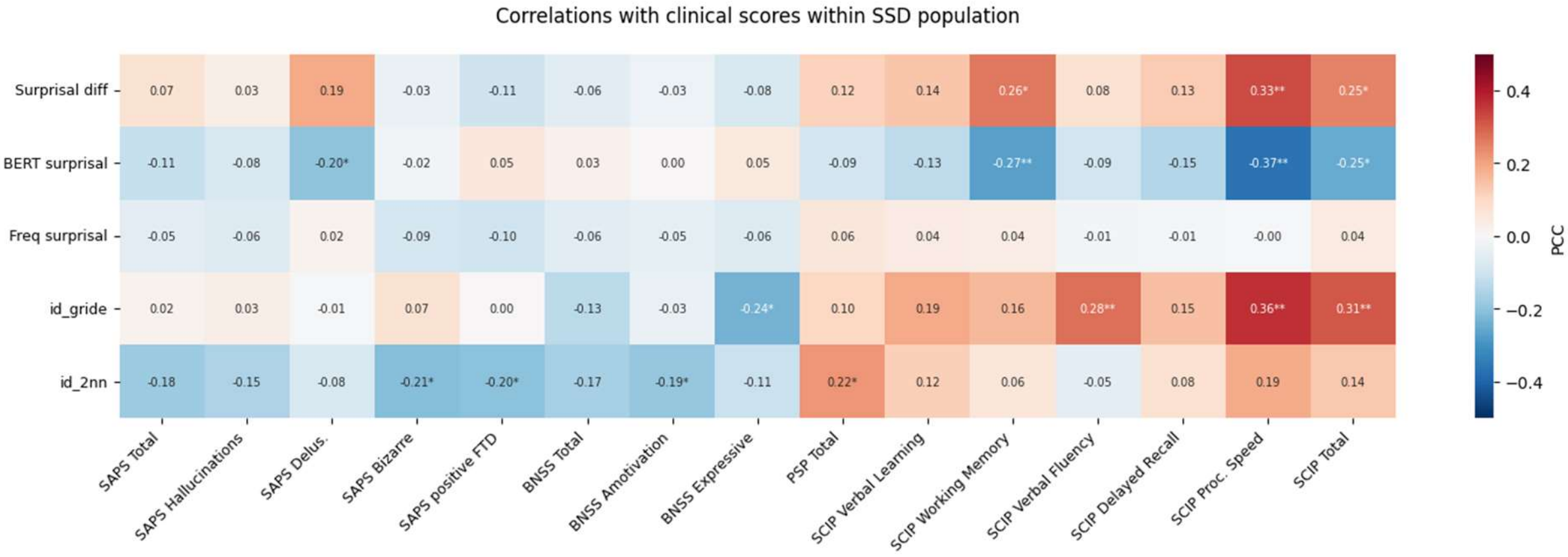

**Figure 4**: Correlations between surprisal and dimensionality measures and clinical scores in patients in SSD. A '*' signal $p - value < 0.05$, '**' $p - value < 0.01$. Strong positive correlations were found between ID Gride and some of the cognitive scores, while negative correlations between ID 2NN and positive and negative symptoms scores.

## References

1. Hinzen, W. & Palaniyappan, L. The 'L-factor': Language as a transdiagnostic dimension in psychopathology. *Prog. Neuropsychopharmacol. Biol. Psychiatry* **131**, (2024).
2. He, R. *et al.* Navigating the semantic space: Unraveling the structure of meaning in psychosis using different computational language models. *Psychiatry Res.* **333**, (2024).
3. Gutiérrez, E., Quesada, C., DeFraites, E., Harper, D. J. & Mandavia, A. D. Interpretable LLM–Based Detection of Loose Associations Using Synthetic Speech Data in Early Psychosis. *Schizophr. Bull.* https://doi.org/10.1093/schbul/sbaf125 (2025) doi:10.1093/schbul/sbaf125.
4. Sharpe, V. *et al.* Selective Insensitivity to Global Versus Local Linguistic Context in Speech Produced by Patients With Untreated Psychosis and Positive Thought Disorder. *Biol. Psychiatry* https://doi.org/10.1016/j.biopsych.2025.06.001 (2025) doi:10.1016/j.biopsych.2025.06.001.
5. Sharpe, V., Weber, K. & Kuperberg, G. R. Impairments in Probabilistic Prediction and Bayesian Learning Can Explain Reduced Neural Semantic Priming in Schizophrenia. *Schizophr. Bull.* **46**, 1558–1566 (2020).
6. Çabuk, T. *et al.* Natural language processing for defining linguistic features in schizophrenia: A sample from Turkish speakers. *Schizophr. Res.* **266**, 183–189 (2024).
7. Voppel, A. E., de Boer, J. N., Brederoo, S. G., Schnack, H. G. & Sommer, I. E. C. Quantified language connectedness in schizophrenia-spectrum disorders. *Psychiatry Res.* **304**, (2021).
8. Arslan, B. *et al.* Automated linguistic analysis in speech samples of Turkish-speaking patients with schizophrenia-spectrum disorders. *Schizophr. Res.* **267**, 65–71 (2024).

9. Palominos, C. *et al.* Approximating the semantic space: word embedding techniques in psychiatric speech analysis. *Schizophrenia* **10**, (2024).
10. Palominos, C. *et al.* Lexical meaning is lower dimensional in psychosis. *Sci. Rep.* https://doi.org/10.1038/s41598-025-30443-1 (2025) doi:10.1038/s41598-025-30443-1.
11. Shannon, C. E. A Mathematical Theory of Communication. *The Bell System Technical Journal* **27**, 379–423 (1948).
12. Viswanathan, K., Gardinazzi, Y., Panerai, G., Cazzaniga, A. & Biagetti, M. The Geometry of Tokens in Internal Representations of Large Language Models. http://arxiv.org/abs/2501.10573 (2025).
13. He, R. *et al. The Grip of Grammar on Meaning Uncertainty: Cross-Linguistic Evidence, Neural Correlates, and Clinical Relevance*. Preprint at https://arxiv.org/abs/2605.01537 (2026)
14. Sharpe, V. *et al.* Selective Insensitivity to Global Versus Local Linguistic Context in Speech Produced by Patients With Untreated Psychosis and Positive Thought Disorder. *Biol. Psychiatry* **99**, 154–164 (2026).
15. Sharpe, V., Weber, K. & Kuperberg, G. R. Impairments in Probabilistic Prediction and Bayesian Learning Can Explain Reduced Neural Semantic Priming in Schizophrenia. *Schizophr. Bull.* **46**, 1558–1566 (2020).
16. Palominos, C. *et al.* Lexical meaning is lower dimensional in psychosis. *Sci. Rep.* **16**, (2026).
17. Speer, R. rspeer/wordfreq: v3.0. Preprint at (2022).
18. Liu, H., Xu, C. & Liang, J. Dependency distance: A new perspective on syntactic patterns in natural languages. *Physics of Life Reviews* vol. 21 171–193 Preprint at https://doi.org/10.1016/j.plrev.2017.03.002 (2017).
19. Ferrer-I-Cancho, R., Gómez-Rodríguez, C., Esteban, J. L. & Alemany-Puig, L. Optimality of syntactic dependency distances. *Phys. Rev. E* **105**, (2022).
20. Devlin, J., Chang, M.-W., Lee, K., Google, K. T. & Language, A. I. *BERT: Pre-Training of Deep Bidirectional Transformers for Language Understanding*. https://github.com/tensorflow/tensor2tensor.
21. Bennett, R. The intrinsic dimensionality of signal collections. *IEEE Trans. Inf. Theor.* **15**, 517–525 (2006).
22. Amsaleg, L. *et al.* Estimating local intrinsic dimensionality. in *Proceedings of the ACM SIGKDD International Conference on Knowledge Discovery and Data Mining* vols 2015-August 29–38 (Association for Computing Machinery, 2015).
23. Facco, E., D'Errico, M., Rodriguez, A. & Laio, A. Estimating the intrinsic dimension of datasets by a minimal neighborhood information. *Sci. Rep.* **7**, (2017).
24. Allegra, M., Facco, E., Denti, F., Laio, A. & Mira, A. Data segmentation based on the local intrinsic dimension. *Sci. Rep.* **10**, (2020).
25. Denti, F., Doimo, D., Laio, A. & Mira, A. Distributional Results for Model-Based Intrinsic Dimension Estimators. http://arxiv.org/abs/2104.13832 (2021).
26. Bac, J., Mirkes, E. M., Gorban, A. N., Tyukin, I. & Zinovyev, A. Scikit-dimension: A python package for intrinsic dimension estimation. *Entropy* **23**, (2021).
27. Denti, F., Doimo, D., Laio, A. & Mira, A. The generalized ratios intrinsic dimension estimator. *Sci. Rep.* **12**, (2022).
28. Denti, F., Doimo, D., Laio, A. & Mira, A. Distributional Results for Model-Based Intrinsic Dimension Estimators. http://arxiv.org/abs/2104.13832 (2021).
29. Glielmo, A. *et al.* DADApy: Distance-based analysis of data-manifolds in Python. *Patterns* **3**, (2022).
30. Vasic, J. *et al.* Lyapunov Spectral Analysis of Speech Embedding Trajectories in Psychosis. http://arxiv.org/abs/2602.16273 (2026).

31. Rosillo-Rodes, P., Miguel, M. S. & Sanchez, D. Entropy and type-token ratio in gigaword corpora. https://doi.org/10.1103/rxxz-lk3n (2025) doi:10.1103/rxxz-lk3n.
32. Xia, M. *et al. Training Trajectories of Language Models Across Scales*. vol. 1 https://github.com/.
33. Pedashenko, V. *et al.* Unveiling Intrinsic Dimension of Texts: from Academic Abstract to Creative Story. http://arxiv.org/abs/2511.15210 (2025).
34. Bergam, N. & Blumberg, A. J. *On Manifold Dimension Estimation*. (2024).
35. Di Noia, A., Macocco, I., Glielmo, A., Laio, A. & Mira, A. *Scale Adaptive and Robust Intrinsic Dimension Estimation via Optimal Neighbourhood Identification ACCEPTED MANUSCRIPT*. (2026).
36. Jazayeri, M. & Ostojic, S. *Interpreting Neural Computations by Examining Intrinsic and Embedding Dimensionality of Neural Activity*.
37. He, R. *et al.* Reduced linguistic coherence in psychosis defies semantic similarity accounts and relates to altered large-scale cortical hierarchy. *Sci. Rep.* **16**, (2026).
38. Di Noia, A., Macocco, I., Glielmo, A., Laio, A. & Mira, A. Beyond the noise: intrinsic dimension estimation with optimal neighbourhood identification. http://arxiv.org/abs/2405.15132 (2026).
39. Di Noia, A., Ravenda, F. & Mira, A. A general framework for adaptive nonparametric dimensionality reduction. *Sci. Rep.* **16**, 9028 (2026).